\documentclass[letterpaper, 11 pt, onecolumn]{ieeeconf}  

\IEEEoverridecommandlockouts                              

\usepackage{graphicx}

\usepackage{cite}

\title{\LARGE \bf
Morphological and Embedded Computation in a Self-contained Soft Robotic Hand}

\author{Nicholas Farrow, Yang Li and Nikolaus Correll
\thanks{This work was supported by NSF NRI grant \#1225934. We are grateful for this support. }
\thanks{All authors are with the Department of Computer Science, University of Colorado, Boulder, 80309, USA
        {\tt\small firstname.lastname@colorado.edu}}%
}

\begin{document}

\maketitle
\thispagestyle{empty}
\pagestyle{empty}

\begin{abstract}
We present a self-contained, soft robotic hand composed of soft pneumatic actuator modules that are equipped with strain and pressure sensing. We show how this data can be used to discern whether a grasp was successful. Co-locating sensing and embedded computation with the actuators greatly simplifies control and system integration. Equipped with a small pump, the hand is self-contained and needs only power and data supplied by a single USB connection to a PC. We demonstrate its function by grasping a variety of objects ranging from very small to large and heavy objects weighing more than the hand itself. The presented system nicely illustrates the advantages of soft robotics: low cost, low weight, and intrinsic compliance. We exploit morphological computation to simplify control, which allows successful grasping via underactuation. Grasping indeed relies on morphological computation at multiple levels, ranging from the geometry of the actuator which  determines the actuator's kinematics, embedded strain sensors to measure curvature, to maximizing contact area and applied force during grasping. 
Morphological computation reaches its limitations, however, when objects are too bulky to self-align with the gripper or when the state of grasping is of interest. We therefore argue that efficient and reliable grasping also requires not only intrinsic compliance, but also embedded sensing and computation. In particular, we show how embedded sensing can be used to detect successful grasps and vary the force exerted onto an object based on local feedback, which is not possible using morphological computation alone. 
\end{abstract}


\section{Introduction}
Soft robotics is an emerging subfield of robotics, which promises compliant sensors and actuators that are inexpensive to manufacture, safe, and allow exploiting morphological computation \cite{pfeifer2006morphological} to off-load challenges that arise during grasping and locomotion into the material design process \cite{rus2015design}. Recent breakthroughs such as the demonstration of reliable grasping using a simple inflatable gripper \cite{ilievski2011soft}, the ability to tune the kinematics of the resulting structures by smart patterning \cite{martinez2012elastomeric}, as well as their integration into robotic systems \cite{correll10iser}, has renewed interest in actuation via inflation. Yet, soft robotics still poses major challenges that prevent this technology from mainstream adaptation and industrial use.

The advantages and drawbacks of soft robotics are nicely illustrated by the McKibben actuator which was developed in the 1950ies, and is one of the more prominent ``pneumatic artificial muscles''\cite{daerden2002pneumatic}. Comprised of a simple inflating membrane constrained by an extensible mesh, this actuator is light-weight, cheap, compliant and very strong. Despite these obvious advantages, its practical uses in robotics are minimal at best. We believe this is primarily due to the difficulty of controlling such a device precisely and accurately. Control is indeed difficult as the actuator has strongly non-linear characteristics which are rooted both in the expansion and friction characteristics of the flexible membrane and the constraint layer. At the same time, proprioception is only of limited use as the soft actuator's mechanical constraints are flexible. This is in contrast with more conventional robotic actuators such as servo motors, which have highly accurate kinematic constraints, and provide precise digital feedback via encoders and measures of current consumption. 
 
There exist multiple ways to simplify the control problem of a soft robot. One is to constrain the design of the actuator so that the control problem becomes more tractable. This could be achieved by patterning that determines the direction of shape change during inflation \cite{ilievski2011soft}, altering the geometry or the rubber characteristics to make the behavior more linear, or even use mechanical solutions such as pressure relief valves, e.g., to implement simple feedback control in the mechanism. These approaches can be broadly categorized as ``morphological computation'' as they simplify the actuator control by exploiting the morphology of the actuator\cite{pfeifer2006morphological}. A complementary approach is to integrate traditional computation more tightly with sensing and actuation \cite{mcevoy2015materials,correll10iser}, thereby abstracting the pneumatic actuator into a device that can receive simple position or force control inputs.

On the other hand, one could argue that soft robotics does not require the precise and accurate control that is central to conventional robotics since the system itself conforms to the task using intrinsic compliance. For example, during grasping it may not be necessary to accurately control each individual finger, as the compliant soft material will deform in a way that increases the effective area of contact forces. In practice however, this is not always sufficient as exploiting environmental constraints such as a table edge or a bowl in order to get an object into the desirable pose is sometimes also required\cite{deimel2014novel}. Here, integrated sensing enables the robot to detect desirable grasp conformations and then move accordingly to maximize the effect of compliance.

\begin{figure}[!htb]
\centering
\includegraphics[width=0.51\columnwidth]{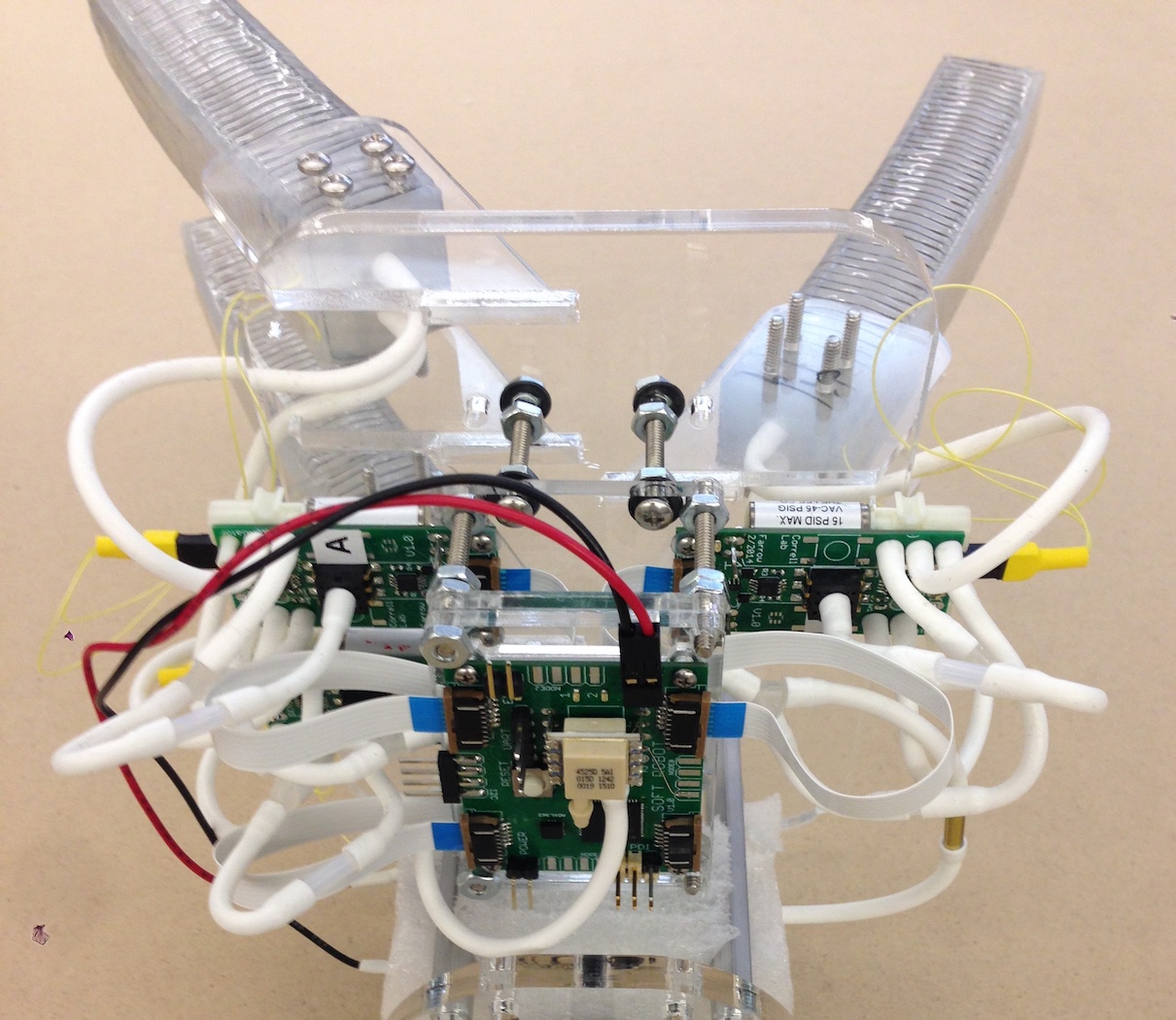}
\includegraphics[width=0.35\columnwidth]{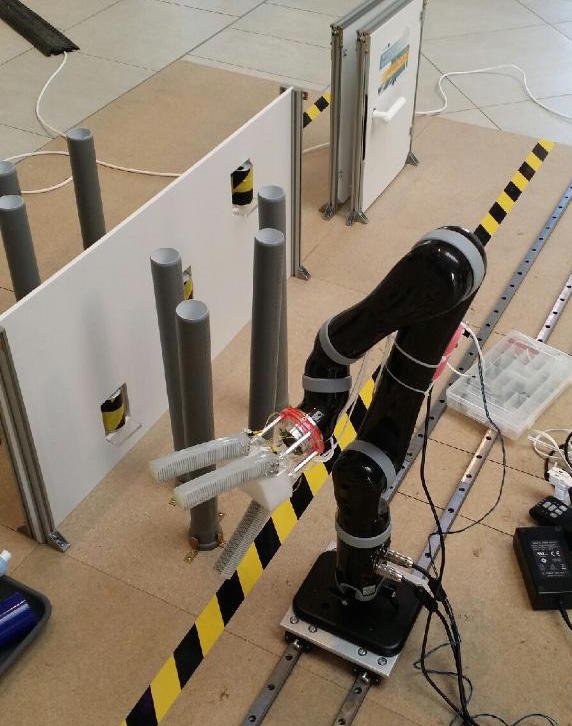}
\caption{Left: A closeup image of the hand assembly seen from behind the wrist. Each finger is controlled by a dedicated control board, which are connected to a central interface. An airpump is hidden behind the central control board and makes the hand self-contained, only requiring a USB connection for power and exchanging sensing and controls with a host system. Right: The hand mounted on a Kinova Jaco light-weight arm at the 2016 RoboSoft Manipulation competition (1$^{st}$ place). While this design combines the advantages of soft, compliant manipulation with the long range afforded by a stiff mechanism, the system was not able to squeeze through the PVC pipes, motivating further research in variable stiffness systems.}
\label{fig:closeup}\label{fig:hand}
\end{figure}

In this paper, we describe our work toward a soft pneumatic actuator (SPA) with integrated strain and pressure sensing, actuated with miniature control electronics, valves, and miniature air compressor. The integrated sensors allow the actuators to autonomously servo to specific curvatures or pressures while sensing internal strain \cite{farrow2015}. At the same time, this actuator maintains the beneficial characteristics of soft robotics. We demonstrate how such SPAs are combined into a more complex system forming a robotic hand (Figure \ref{fig:hand}), and show how the system combines the advantages of morphological computation with conventional sensing and control. 

\section{Related work}

Morphological computation \cite{pfeifer2006morphological} refers to the role of shape and physical first principles in mechanism design to simplify the required computation for signal processing and control. As materials can be designed to exhibit specific non-linear responses that can be combined with each other, morphological computation is theoretically capable of universal computation \cite{hauser2011towards}. In this paper, we refer to morphological computation as the use of form and elasticity to simplify controller design. 

Compliant actuators have been identified as simple solutions for grasping objects with unknown or challenging geometry, leading to a series of hand designs exploiting that concept \cite{deimel2014novel} as well as under-actuated grasping mechanisms \cite{dollar2010highly,amend2012positive,ilievski2011soft,giannaccini2014variable}, which are surprisingly capable. Here, soft pneumatic mechanisms \cite{ilievski2011soft,Kota2012,deimel2014novel,giannaccini2014variable} are particularly attractive as they are compliant and cheap to manufacture. Compliant actuators, while beneficial for grasping, have the disadvantage of being both mathematically and computationally difficult to model. This is because the dynamics of soft materials are much more complicated than the kinematics of a stiff linkage causing them to be difficult to control precisely or model accurately. Further, proprioception in soft robotics has been relatively unexplored so far as identifying simple sensors suitable for integration in deformable materials remains a challenge. In \cite{deimel2014novel,RBOhand2013,ilievski2011soft} actuators are controlled by a simple control without feedback and rely on the compliant nature of the actuators in the closed state to regulate their own position around the object. In \cite{Marchese2014}, a motion capture system is used to track the progress of inflation to perform closed-loop control of a pneumatic arm. In \cite{correll10iser}, simple light sensors are used to control inflation of a pneumatic actuator by exploiting the morphology of the robot and its environment. 
Curvature sensors are explored in \cite{bilodeau15} and \cite{homberg15}. In \cite{bilodeau15} a simple gripper \cite{ilievski2011soft} is equipped with a liquid metal strain gauge with emphasis on the monolithic manufacturing process. In \cite{homberg15}, strain sensors are used in a very similar configuration to those presented here, and used to classify objects the robot is holding rather than to provide feedback on grasp quality.

\section{Design}

We address the limitations that come from an open-loop control approach by incorporating a pair of sensors into the actuator which provide a proprioception mechanism for tighter control of bending and estimating of grasp success. We describe two sensor technologies coupled with each actuator: air pressure sensors and strain sensors composed of liquid metal alloy inspired by \cite{YLPark-3axis-eGaIn2012}. When combined, these sensors provide feedback which enables more accurate control over the positioning of the actuator in both free space and while grasping objects. In particular, we are able to measure not only the curvature of the actuator, but also the pressure required to maintain this position. 

\begin{figure}[thb]
\centering
\includegraphics[width=0.38\columnwidth]{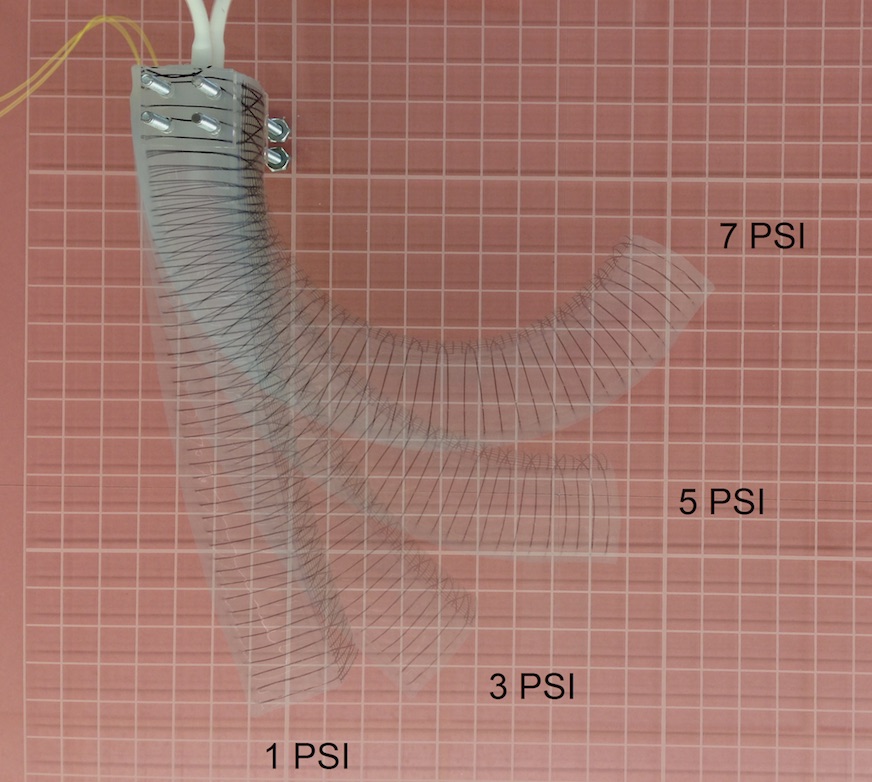}
\includegraphics[width=3.6 in]{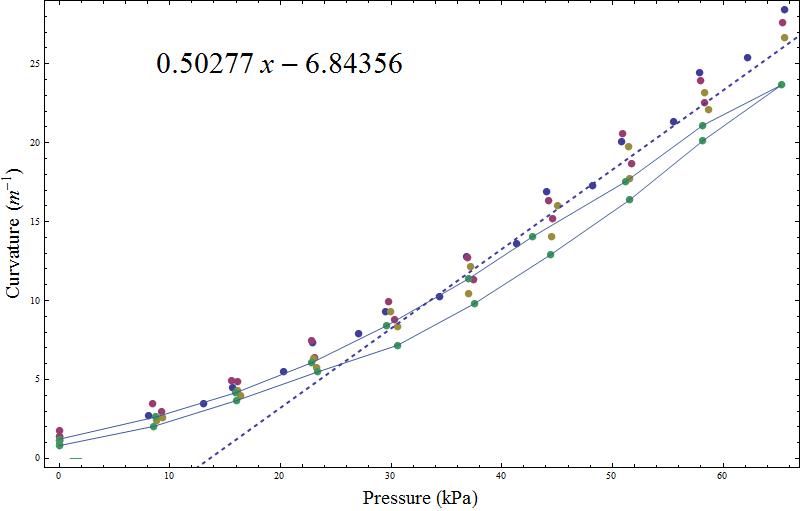}

\caption{Left: A sequence of images of the actuator taken as part of calibration experiments, showing the uniform curvature of the actuator at various applied pressures on a centimeter-scale grid. Right: Measured curvature vs. supplied pressure for four actuators with identical shape, size and wall thicknesses. A solid line is used to connect the data points associated with the one of the actuators, and is intended to show the small hysteresis observed (plots are counterclockwise in time). A dashed line indicates the calibrated line of best fit taken from all of the actuators, above the nominal threshold pressure of 30 kPa (4.35 PSI) that is roughly equivalent to a radius of curvature of 1 meter. 
\label{fig:PressureSequence}\label{fig:PvsK}}
\end{figure}

\begin{figure}[!htb]
\centering
\includegraphics[width=0.74\columnwidth]{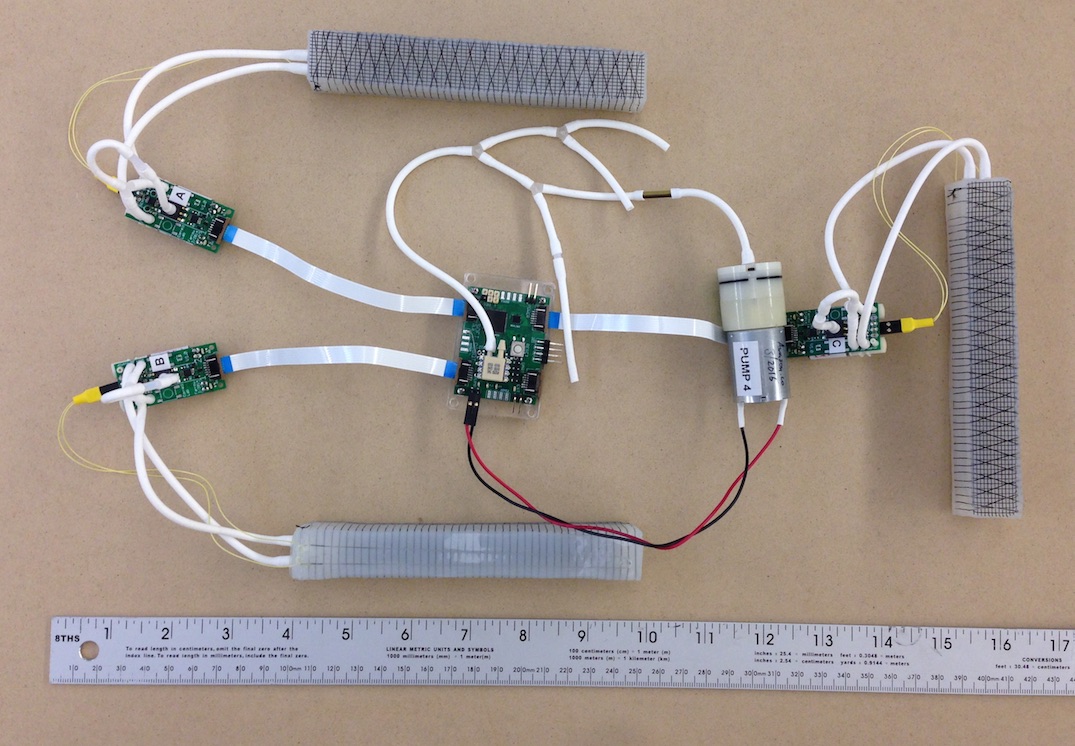}
\includegraphics[width=0.13\columnwidth]{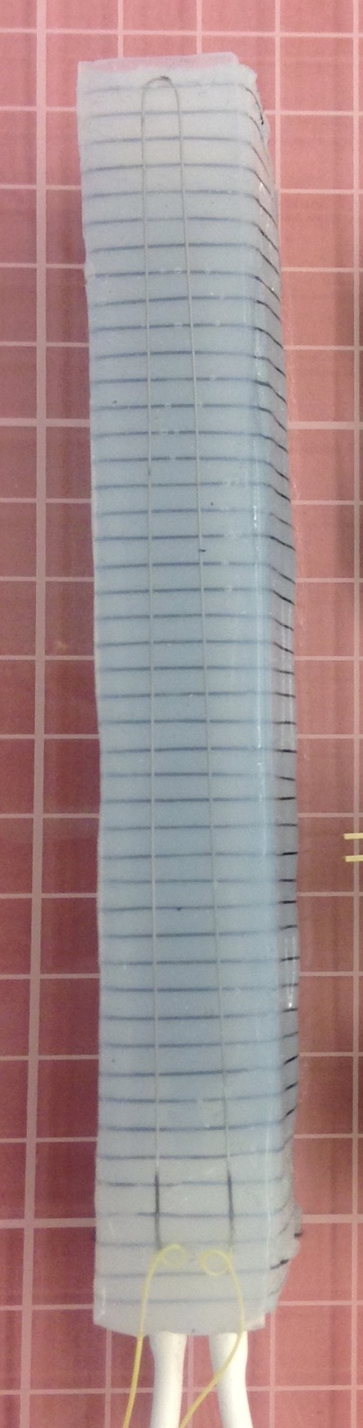}
\caption{Left: A layout of the hardware used in the self-contained hand, including main control PCB (center) surrounded by three reinforced pneumatic actuators and their sensor PCBs. Right: Close-up on a single actuator with integrated sensor \cite{farrow2015}.
\label{fig:actuator}\label{fig:sensor_closeup}}
\end{figure}

Our actuator design builds upon the fiber reinforced actuators described in \cite{RBOhand2013}. Its manufacturing is described in complete detail in \cite{farrow2015}. These actuators consist of a simple elastomer structure (Dragon Skin 10 Medium) enclosing a single rectangular air chamber. Embedded fibers wrapped around the length of the actuator as well as down the ventral surface constrain the inflation both radially and lengthwise along one side, resulting in a bending/curling motion when air pressure is supplied (Figure \ref{fig:PressureSequence}). This internal strain causes bending which is measured using an integrated strain sensor on the dorsal surface of the actuator (Figure \ref{fig:sensor_closeup}, right). Given a calibrated model of the actuator, we can use sensor information to bend our actuator to any feasible curvature, e.g., during grasping, by inflating to a desired pressure. When there is an object in the path of the grasping actuator, the compliant actuator will grip the obstacle with increasing contact force instead of continuing to bend.

We simplify modeling by approximating a linear curvature function of pressure, which is reasonable for moderate bending angles (Figure \ref{fig:PvsK}).  
This model makes the following simplifying assumptions:
The Young's modulus of Dragon Skin 10 is unknown and likely varies from batch to batch \cite{case2015soft}. It is also rate dependent and non-linear as the material is viscoelastic.  We assume that actuation is slow enough that steady state conditions approximate our system.  This approximation is reasonable based on the actuation speeds we observe and considering that the final grasp is always at dynamic equilibrium.
We also note that interactions with the environment, including grasping objects, impose external forces on the system which are not entirely predictable nor included in any analytical or computerized solver.

Another drawback of actuators with rectangular cross-section is that the walls of the actuator deform radially during the initial phase of actuation.  Fibers prevent the actuator from ballooning, but not from deforming, and so the initial deformation with applied pressure causes the cross section of the actuator to go from square to round before bending becomes significant. We presume that this is a primary source of the non-linearity that our actuators exhibit.  In practice, the actuators need only to be inflated above a certain threshold pressure at which deformation occurs and before bending becomes noticeable.  Here, the threshold pressure is approximately 1 psi, or about 10\% of the maximal inflation pressure. We note that the threshold pressure seems to depend on the wall thickness of the actuator, with thicker-walled actuators being less able to deform into circles.

To experimentally evalute the relationship between pressure and curvature, the actuators were anchored horizontally in free space, above a 1-cm grid table surface and inflated to a specific pressure (Figure \ref{fig:PvsK}, left).  Images of the actuators were captured and processed with custom image processing software to compute the radius of each actuator as a function of pressure.  Four actuators, manufactured independently, from different batches of elastomer yet identical in design were each calibrated using the same procedure.  Figure \ref{fig:PvsK}, right, uses a solid line to show the linear approximation that is taken from all the pressures above around 30 kPa. We fully inflated each actuator at least 10 times before calibration, so that the elastomer does not exhibit the Mullins effect \cite{case2015soft}.

\label{sec:sensor}

As part of actuator assembly, we manufacture our own strain sensors very similar to the sensors in \cite{YLPark-3axis-eGaIn2012}, however we deviate in the manufacturing process to avoid a difficult monolithic fabrication technique. Instead, we use pre-fabricated small diameter silicone tubing which we lay out into simple small channels and fix them into place with a very thin elastomer poured on top.  From here, the embedded tubing is cut to size, the exposed ends are capped with more elastomer, and finally the tubes are injected with conductive liquid (Galinstan) to finish the resistive strain sensor.  We affix this very thin strain sensor to the dorsal side of the actuator, which adds about 2 mm to the height of the actuator and increases the actuator stiffness by a marginal amount. The manufacturing process is described in more detail in \cite{farrow2015}. 

We consider the resistive strain sensor itself to be a morphological computer, as it is the morphology of the sensor with respect to shape, pattern and placement that affects how different deformations lead to changes in resistivity.
A miniature instrument amplifier IC (AD8327) on the actuator PCB provides a voltage measurable by our microcontroller's analog-to-digital converter. We have measured the resistance vs.\ strain for four sensors that we manufactured, with a small deviation of the theoretical prediction resulting from the finite lead length of the copper wires connecting the Galinstan strain gauge to the amplifier. The same circuit board also houses a small pressure sensor (US9111) that can measure the pressure inside the actuator (0 to 15 PSI), as well as valves (The Lee Corporation) needed for control (Figure \ref{fig:actuator}).  The main PCB also uses a secondary, digital pressure sensor (MS4525D0) which is a differential pressure sensor.  We calibrate all of the analog sensors on the actuators to the main differential pressure sensor as atmospheric pressure changes noticeably during the day.

\begin{figure}[!htb]
\centering
\includegraphics[width=3.2 in]{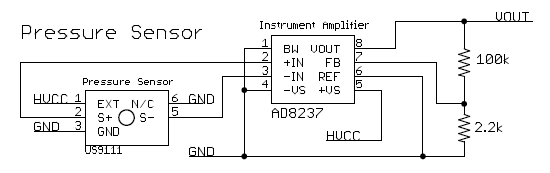}
\includegraphics[width=3 in]{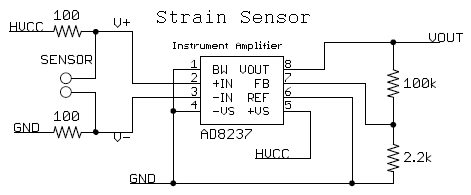}
\caption{Left: The pressure sensor circuit used with this actuator. Right: The strain sensor circuit used with this actuator. This circuit could be adapted to any resistor-based sensor, such as a flex sensor instead.
\label{fig:PressureSensorCircuit}\label{fig:StretchSensorCircuit}}
\end{figure}

Figure \ref{fig:PressureSensorCircuit}, left, shows the schematics of the pressure sensor circuit. The digital pressure sensor used (US9111) maps 0 to 15 PSI to 0 to 100mV, requiring an instrument amplifier (AD8237) before processing by the controller. Figure \ref{fig:StretchSensorCircuit}, right, shows the schematics of the curvature sensor. As the resistance of the Galinstan channel is very low, we are using two 100 $\Omega$ resistors to limit current through the sensor. The instrument amplifier (AD8237) then measures the voltage across the sensor using the same circuit as for the pressure sensor. 

\begin{figure}[!htb]
\centering
\includegraphics[width=0.6\columnwidth]{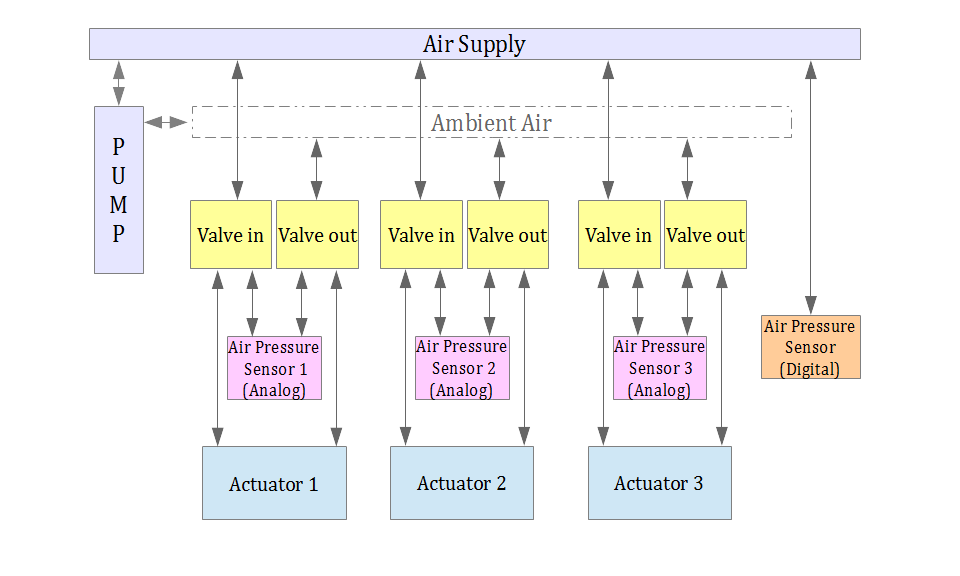} \\
\includegraphics[width=0.4\columnwidth]{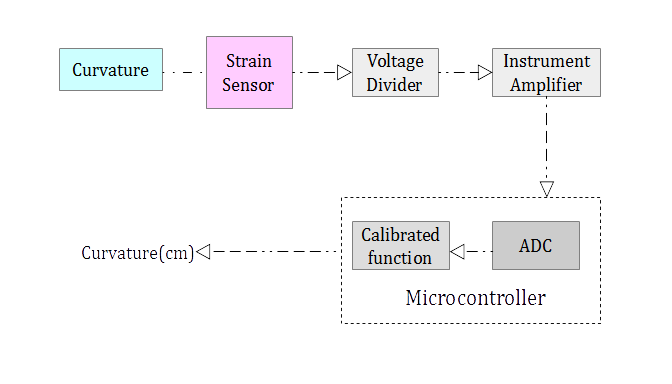}
\includegraphics[width=0.4\columnwidth]{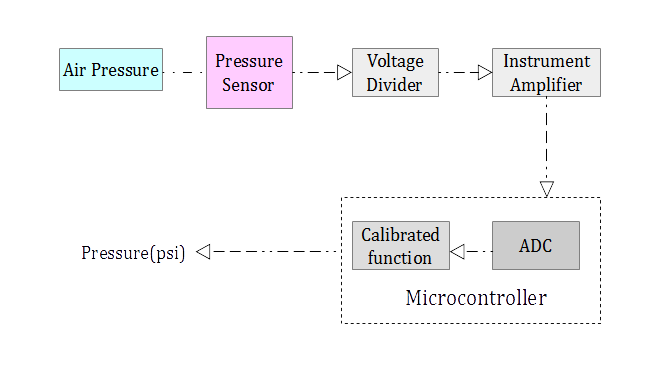}
\caption{Top: The pneumatic circuit connecting the components of the hand. Bottom: The analog measurement pipeline for sensing curvature (left) and air pressure (right) within the actuator, to convert this signal from the sensors into natural units at the microcontroller.
\label{fig:pneumatics}\label{fig:ADCpipeline}}
\end{figure}

The PCB housing the sensing circuitry and valves interfaces to another PCB with a microcontroller (Atmel Atxmega128A3U) that implements all  control, sensing via ADC, and actuation while interfacing to a PC via USB to receive high level commands and to output data.
The control law is straightforward, turning the valve on or off until a certain curvature or pressure level is reached. A high-level finite state machine implemented for each actuator allows the microcontroller to maintain control over one or more actuators (up to six) at the same time via a serial interface.  

\section{System integration and experimental validation}
We constructed a simple hand from laser-cut acrylic sheets that allows us to arrange three actuators in a claw configuration similar to \cite{homberg15} and attached it to the Baxter robot. The hand connects via aluminum stand-offs to a base plate that interfaces Baxter's wrist, leaving space for the main control board and a small air pump typically used with aquarium bubblers. The miniature compressor is capable of producing approximately 10 PSI. The hand is mounted such that Baxter's in-wrist camera is still usable. A block of foam serves as a palm, providing additional constraints to the object during caging grasps. Both arrangements, single finger and hand, interface to a control PC via USB and a simple serial interface. We note that the USB connection is the only interface the resulting hand needs, providing both communication and power to the system. The entire hand including our Baxter baseplate weighs 490 g.

We conducted two kinds of experiments using the three-fingered hand. We first examine the sensor response of an individual actuator when grasping a series of cylindrical objects (bottles, cans, tubes) \cite{farrow2015}. In this experiment, an individual finger was placed close to the object. The actuator was then used to grasp the object but was not required to lift the object. Each actuator was controlled to inflate to around 8 psi and hold firmly to the shape of the object. The pressures and applied forces are relatively small and do not deform the objects. The object prevents the actuator from reaching the full curvature that would ordinarily happen with an empty grasp, resulting in an attenuated strain response.

\begin{figure}[!htb]
\centering
\includegraphics[width=0.4\textwidth]{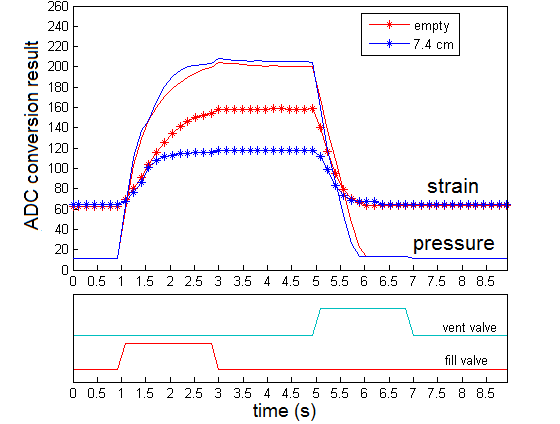}
\includegraphics[width=0.5\textwidth]{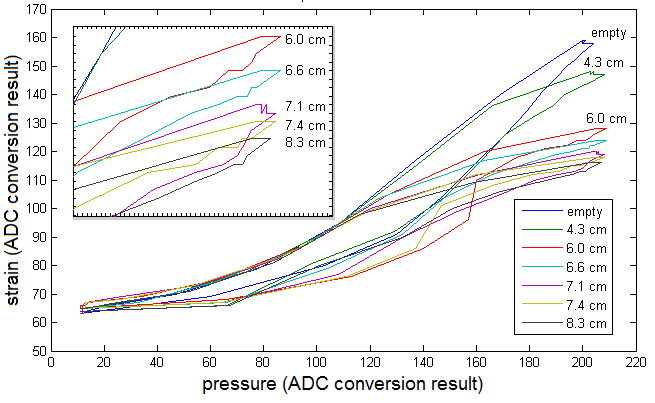}
\caption{Left: Control sequence of grasping. An empty grasp is shown with a successful grasp for comparison. ADC values are proportional to strain sensor readings/curvature. Right: A phase plot of grasping a set of round objects with known diameters. All grasps begin at low pressure and low strain (bottom left of plot). Phase orbits are counterclockwise. Full actuation occurs at the maximal curvature point of the phase orbit. Any deviation from the empty grasp orbit indicates an interaction with the environment.
\label{fig:GraspComparePlot}\label{fig:RoundGraspCalibration}}
\end{figure}

Measurements comparing an empty grasp to a successful grasp of a cylindrical object with 7.4 cm radius together with actuator signals are shown in Figure \ref{fig:GraspComparePlot}, left. The combined sensor readings of strain and pressure are used to determine grasp radius and to estimate the success of the grasp. Indeed, one can clearly observe that pressure keeps changing albeit strain remains constant, suggesting a successful grasp. 
%
A phase plot of the strain vs.\ pressure sensor illustrates the subtleties of the sensing process, and is illustrated in Figure \ref{fig:RoundGraspCalibration}, right. The grasp phase plot shows that the largest strain happens during an empty grasp, while increasingly smaller strains are observed with larger diameter objects. Multiple grasp measurements were made for the objects as well as empty grasps, and the data reveal qualitatively identical phase plots (not shown) indicating that the grasp phase orbits are reliable indicators of grasp success. We believe the hysteresis in the phase plot originates from viscoelastic properties of the material. 

\begin{figure*}[!htb]
\centering
\includegraphics[width=0.25\textwidth]{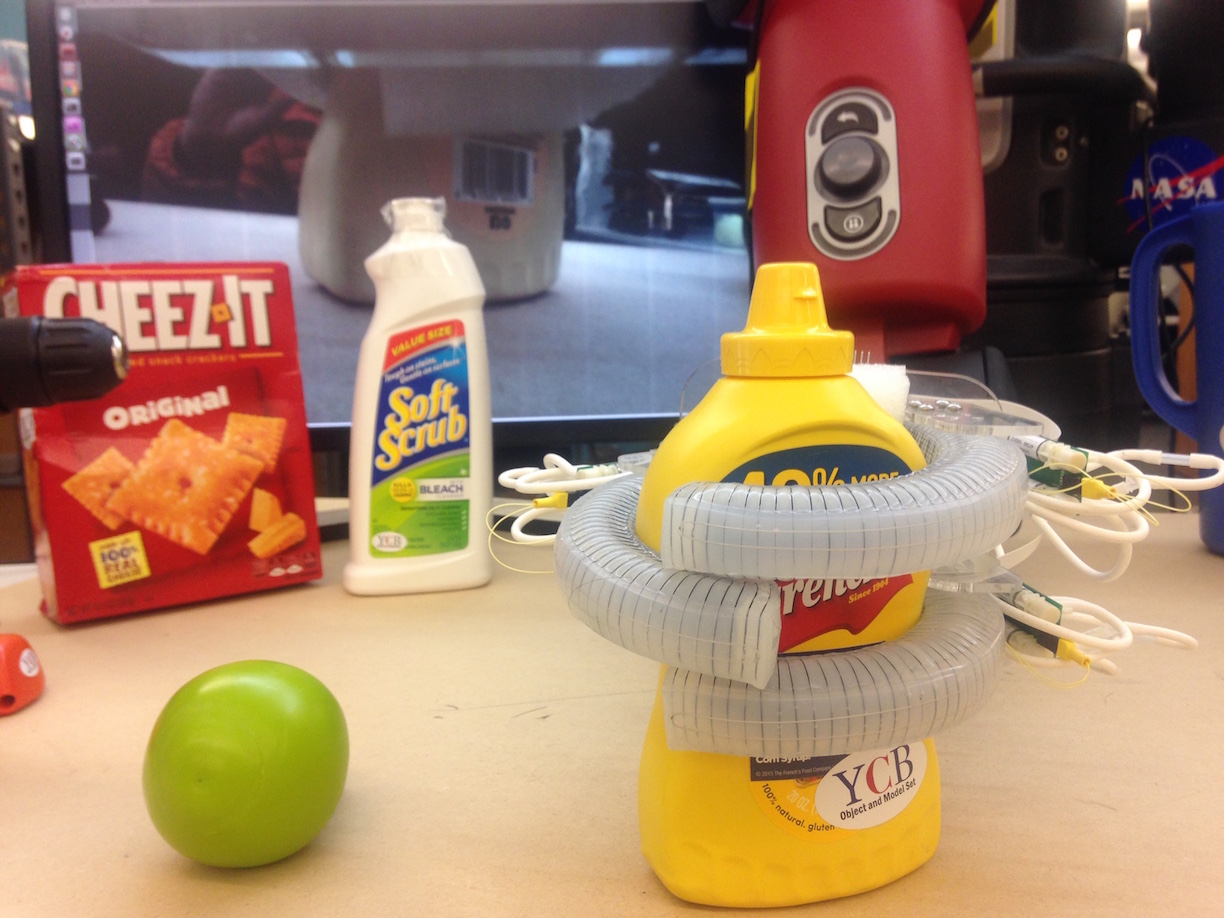}
\includegraphics[width=0.25\textwidth]{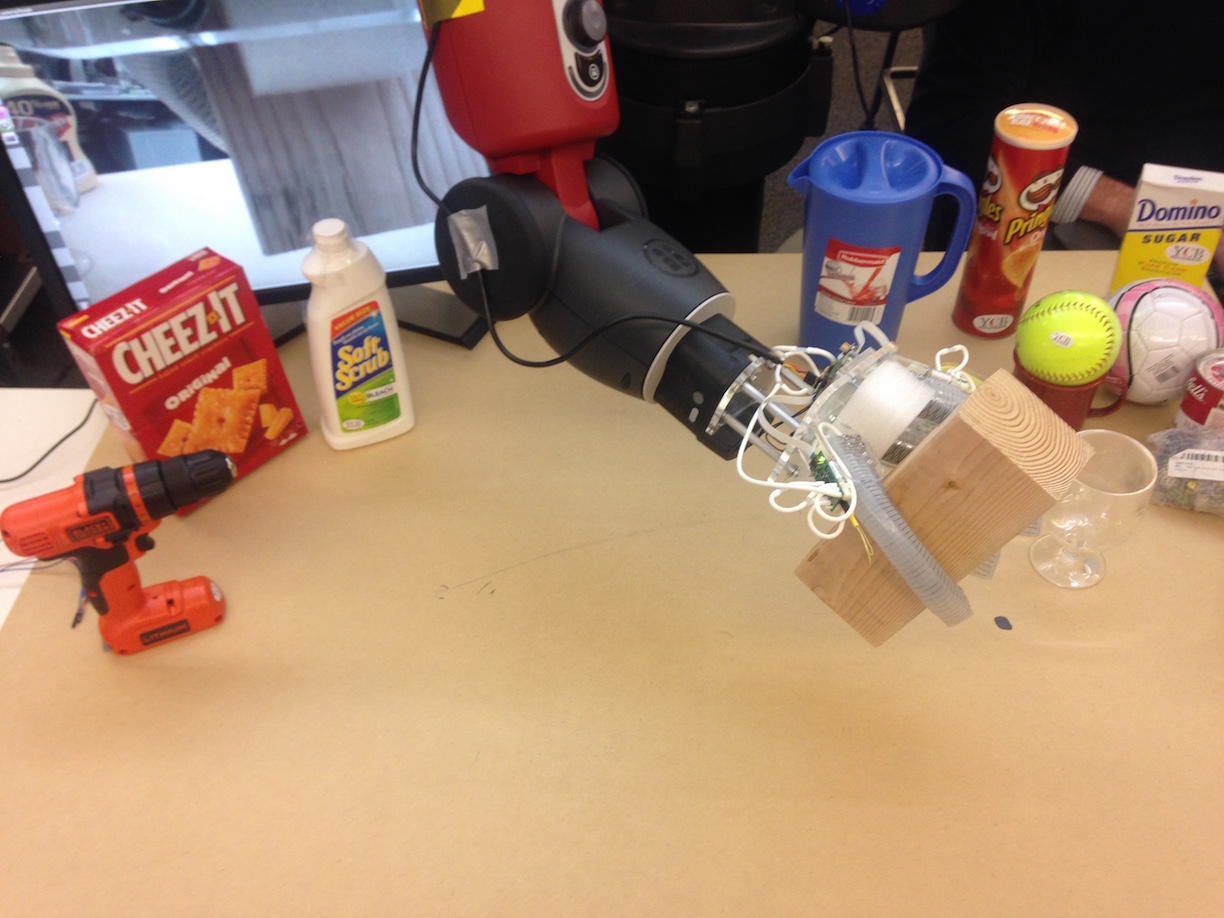}
\includegraphics[width=0.19\textwidth]{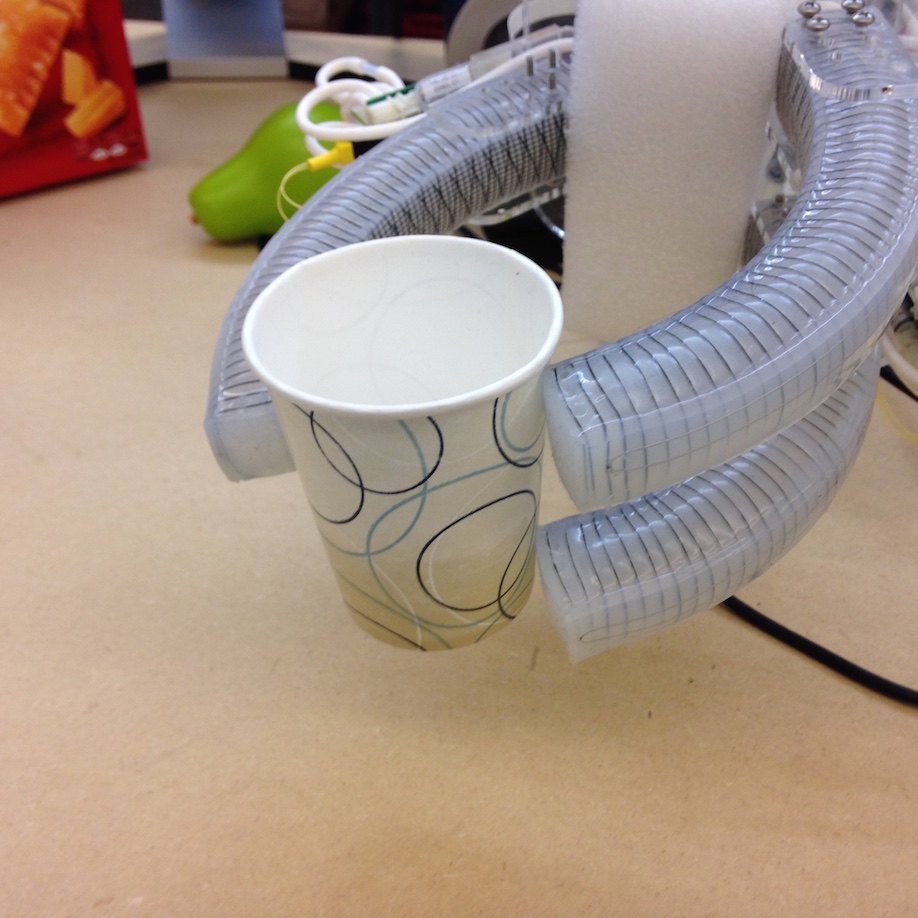}
\includegraphics[width=0.19\textwidth]{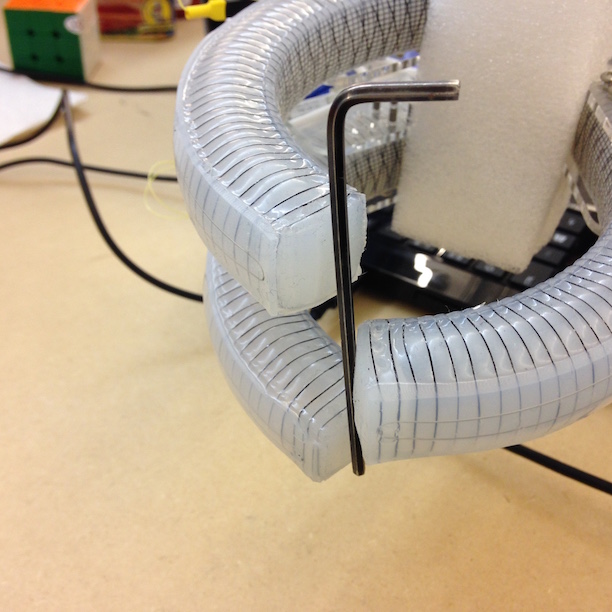}
\caption{Grasping various objects from \cite{calli2015ycb}, including the pitcher, mustard container, and piece of wood. Other objects that can be successfully grasped using this configuration are shown in the background. Small objects can be grasped using a pinching grasp. From left to right: small paper cup, hex key.
\label{fig:ycbgrasps}} 
\end{figure*}

For the second experiment, we manually grasped a series of objects shown in Figure \ref{fig:ycbgrasps} using both caging and pinching grasp. Here, the wooden block from the YCB set that weighs 770g (1.7 lb) pushes the actuators to the limits, which begin bending downwards.
Caging grasps occasionally lead to poor initial force closure, whereas pinching grasps (using the finger tips to grasp) sometimes miss the object. Some grasps did not immediately lead to complete force closure around the object. It is therefore not always possible to obtain good quality grasps on the initial grasp closure using passive grasping alone. In particular, heavier items such as the YCB mustard container (628g) are more difficult to grasp and lift than light-weight large items such as the empty YCB pitcher (244g) as they require maximum force-closure for not slipping through the hand.

\begin{figure}[!htb]
\begin{tabular}{p{0.3\columnwidth} c}
\vspace{-200px}
\includegraphics[width=0.25\columnwidth]{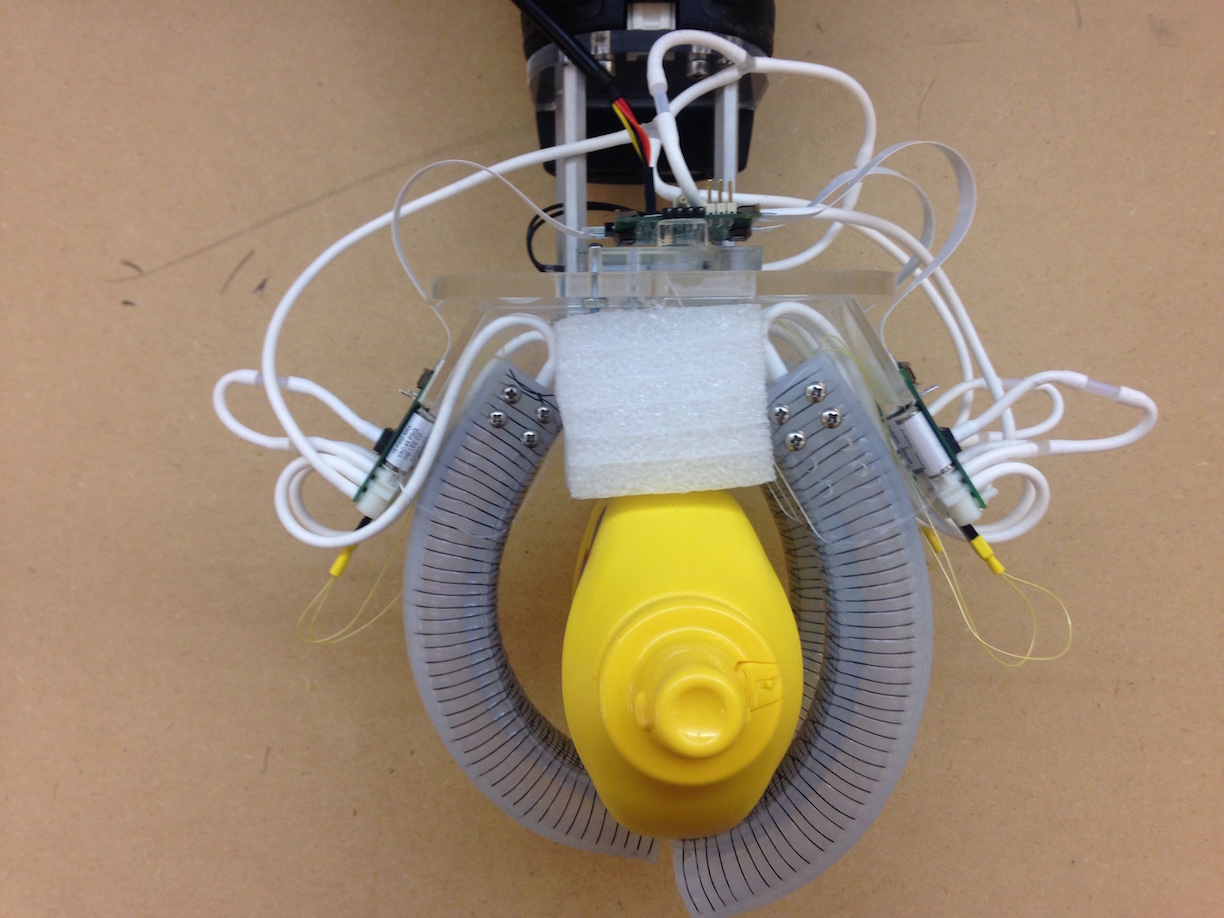}
\includegraphics[width=0.25\columnwidth]{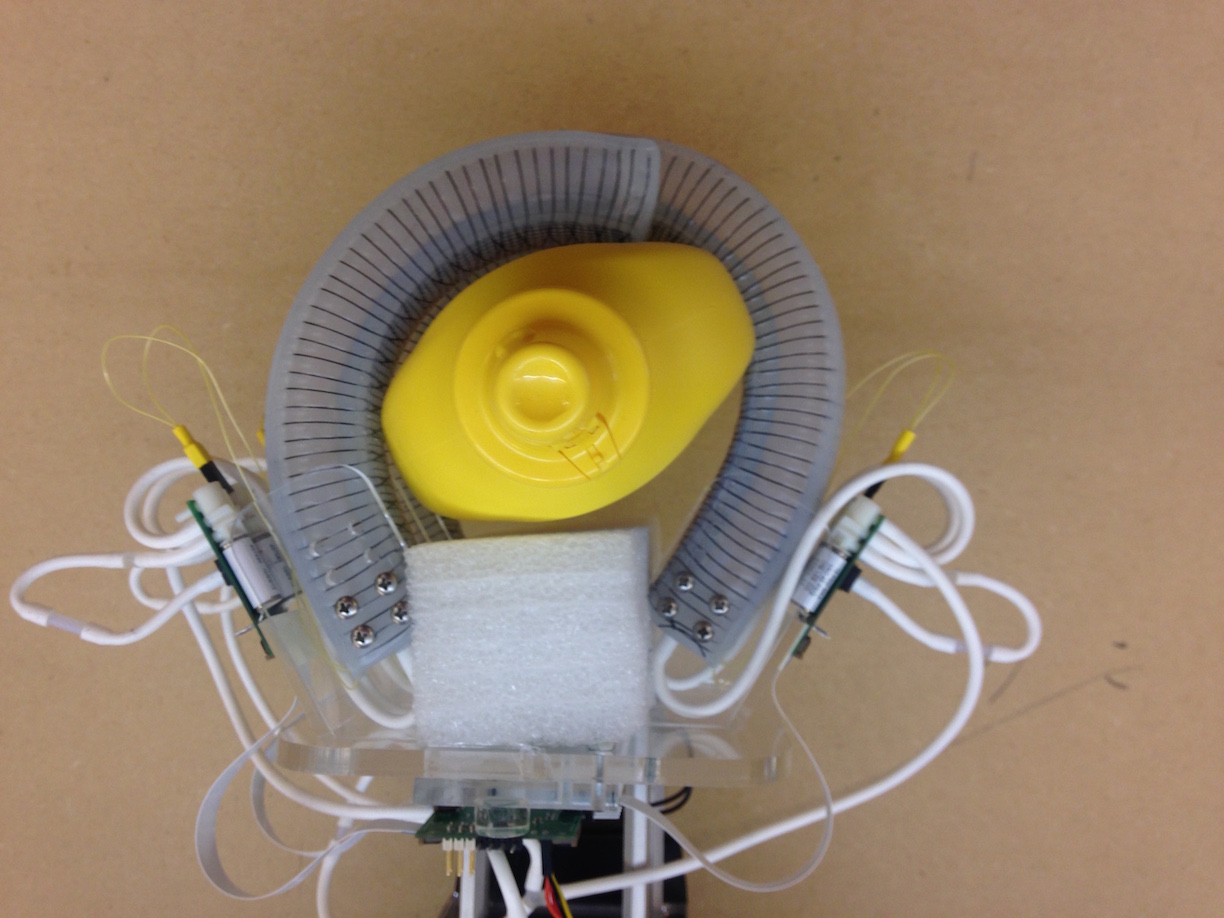}
&
\includegraphics[width=0.6\columnwidth]{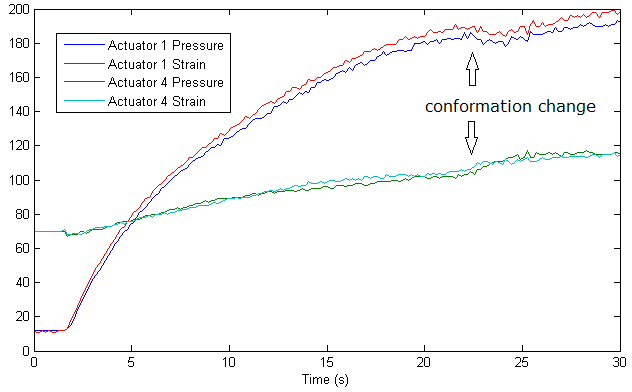}
\end{tabular}
\caption{Initial grasp of the mustard container (top left) with poor force-closure. Improved force-closure after gently wiggling of the wrist (bottom left). Pressure and strain in the left and right actuator vs.\ time. Confirmation changes are visible in abrupt changes of pressure and curvature. The grasp is complete once curvature settles. \label{fig:shakingmustard}}
\end{figure}

We therefore conducted another set of experiments where we first performed a caging grasp on an intentionally poorly aligned object. We then begin to moderately shake and twist the hand. This additional motion allowed both the object and actuators to re-align themselves while significantly increasing force closure. An example grasp of the mustard container before and after gentle shaking is shown in Figure \ref{fig:shakingmustard}. We have also recorded both pressure and strain data from the actuators surrounding the object while wiggling the gripper. Abrupt grasp conformation changes are visible in both pressure and curvature at around second 23. This, together with the ability to differentiate between successful and empty grasps, is an indicator that this information might be useful during autonomous grasping, which we will explore in the future.  

\vspace{25px}

\section{Discussion}

We described a soft robotic hand with individually articulated fingers that is capable of lifting heavy items that exceed the hand's own weight using only the power supplied via a laptop's USB port, demonstrating the potential of soft robotic technology for grasping a range of objects. For comparison, the Yale Open hand's ``Model O'' weighs around 800g, owing most of its weight to the integrated servo motors, and requires up to 6A for its motors to reach their stall torque. A lightweight, simple three-fingered claw design using pneumatic actuators such as presented here is capable to also grasp small and thin items such as pens, small tools or paper (see also \cite{homberg15}).  

We have demonstrated how sensing of curvature and pressure can be used in a feedback control loop, thereby making up for uncertainty in the intrinsic non-linearities of the soft actuator. Morphological computation, that is relying on compliance and smart sensor placement, is used to greatly simplify the controller. While further improving the hardware design may indeed have the potential to improve the controllability, impulse response, and bandwidth of the system, it is unclear how more advanced features that computation affords, such as PID force control or object recognition \cite{homberg15} can be implemented using morphological computation in the sense of \cite{hauser2011towards} without substantially altering the kinematics of the system. 
 
We observe morphological computation to be a powerful aide in ``calculating'' optimal force-closure (Figure \ref{fig:shakingmustard}). Allowing the object to move by wiggling the hand allows the actuators to minimize their internal energy, eventually leading to a conformation that minimizes the potential energy of the object (with respect of degrees of freedom that remain within the claw) and the bending energy of each actuator. Yet, morphological computation alone is not sufficient to determine when to stop wiggling. Instead, more (conventional) computation is required to determine if a conformation change occurs or when a steady state is reached and lifting the object can be attempted. Studying this effect systematically using a large variety of objects is subject to future work. 

Whereas the proposed design is capable of holding a large variety of objects, performing grasping completely autonomously remains unsolved. Even though caging can be used to reliably grasp larger objects, a robot still needs to plan for collision-free paths for all of its fingers, for example when grasping a pitcher through its handle. Performing pinching grasps is quite difficult as the hand needs to be precisely aligned with the object before the fingers are closed. Here, tactile or pre-grasp sensing might be exploited to ``feel'' out an object before making a final grasp, which we wish to explore in the future. 

A drawback of soft actuators are their inability to support larger loads. Although the presented hand is able to support the 750g wooden block from the YCB data set and effect sufficient force to operate a door handle, a stiff mechanism is needed to support the hand, loosing the advantages of compliance for this part of the robot. In future work, we wish to investigate variable stiffness mechanisms \cite{mcevoy2016shape} to extend the advantages of compliance and flexibility to the entire mechanism, which might allow solving hard manipulation problems such as shown in Figure \ref{fig:closeup} that involve squeezing the manipulator through tight openings. 

\section{Conclusion}
Soft robotics enables powerful actuation at low complexity and cost. The resulting systems are simple to manufacture and cheap, relying on simple molded rubber actuators, laser cut acrylic and commodity parts. Compliance of the actuators is not only attractive for safe interactions with humans, but also provides a means to simplify the grasping problem by allowing for underactuated operation. We have also shown, however, that certain tasks cannot reliably be accomplished without additional sensing and computation. In particular, we demonstrate that sensing both curvature and pressure allows to not only limit forces exerted onto an object, but also allows recognizing the state the gripper is in, which is important for assessing the quality of a grasp. 

Miniaturizing sensing, control, and actuation to the point where they are co-located with individual actuators has allowed us to abstract the control problem that is inherent to soft robotics. By moving the burden of sensing and feedback control into the actuator, its interface now resembles that of a digital servo motor. We believe that it is such abstractions that will allow soft robotics to move from a niche application to mainstream use throughout robotics. 

\bibliographystyle{IEEEtran}
\bibliography{IEEEabrv,ram_farrow}

\begin{thebibliography}{10}
\providecommand{\url}[1]{#1}
\csname url@samestyle\endcsname
\providecommand{\newblock}{\relax}
\providecommand{\bibinfo}[2]{#2}
\providecommand{\BIBentrySTDinterwordspacing}{\spaceskip=0pt\relax}
\providecommand{\BIBentryALTinterwordstretchfactor}{4}
\providecommand{\BIBentryALTinterwordspacing}{\spaceskip=\fontdimen2\font plus
\BIBentryALTinterwordstretchfactor\fontdimen3\font minus
  \fontdimen4\font\relax}
\providecommand{\BIBforeignlanguage}[2]{{%
\expandafter\ifx\csname l@#1\endcsname\relax
\typeout{** WARNING: IEEEtran.bst: No hyphenation pattern has been}%
\typeout{** loaded for the language `#1'. Using the pattern for}%
\typeout{** the default language instead.}%
\else
\language=\csname l@#1\endcsname
\fi
#2}}
\providecommand{\BIBdecl}{\relax}
\BIBdecl

\bibitem{pfeifer2006morphological}
R.~Pfeifer, F.~Iida, and G.~G{\'o}mez, ``Morphological computation for adaptive
  behavior and cognition,'' in \emph{International Congress Series}, vol.
  1291.\hskip 1em plus 0.5em minus 0.4em\relax Elsevier, 2006, pp. 22--29.

\bibitem{rus2015design}
D.~Rus and M.~T. Tolley, ``Design, fabrication and control of soft robots,''
  \emph{Nature}, vol. 521, no. 7553, pp. 467--475, 2015.

\bibitem{ilievski2011soft}
F.~Ilievski, A.~D. Mazzeo, R.~F. Shepherd, X.~Chen, and G.~M. Whitesides,
  ``Soft robotics for chemists,'' \emph{Angewandte Chemie}, vol. 123, no.~8,
  pp. 1930--1935, 2011.

\bibitem{martinez2012elastomeric}
R.~V. Martinez, C.~R. Fish, X.~Chen, and G.~M. Whitesides, ``Elastomeric
  origami: Programmable paper-elastomer composites as pneumatic actuators,''
  \emph{Advanced Functional Materials}, vol.~22, no.~7, pp. 1376--1384, 2012.

\bibitem{correll10iser}
N.~Correll, C.~Onal, H.~Liang, E.~Schoenfeld, and D.~Rus, ``Soft autonomous
  materials - using programmed elasticity and embedded distributed
  computation,'' in \emph{International Symposium on Experimental Robotics
  (ISER). Springer Tracts in Advanced Robotics.}, V.~K. Oussama~Kahtib and
  G.~Sukhatme, Eds., 2010.

\bibitem{daerden2002pneumatic}
F.~Daerden and D.~Lefeber, ``Pneumatic artificial muscles: actuators for
  robotics and automation,'' \emph{European journal of mechanical and
  environmental engineering}, vol.~47, no.~1, pp. 11--21, 2002.

\bibitem{mcevoy2015materials}
M.~McEvoy and N.~Correll, ``Materials that couple sensing, actuation,
  computation, and communication,'' \emph{Science}, vol. 347, no. 6228, p.
  1261689, 2015.

\bibitem{deimel2014novel}
R.~Deimel and O.~Brock, ``A novel type of compliant, underactuated robotic hand
  for dexterous grasping,'' \emph{Robotics: Science and Systems, Berkeley, CA},
  pp. 1687--1692, 2014.

\bibitem{farrow2015}
N.~Farrow and N.~Correll, ``A soft pneumatic actuator that can sense grasp and
  touch,'' in \emph{Intelligent Robots and Systems (IROS), 2015 IEEE/RSJ
  International Conference on}.\hskip 1em plus 0.5em minus 0.4em\relax IEEE,
  2015, pp. 2317--2323.

\bibitem{hauser2011towards}
H.~Hauser, A.~J. Ijspeert, R.~M. F{\"u}chslin, R.~Pfeifer, and W.~Maass,
  ``Towards a theoretical foundation for morphological computation with
  compliant bodies,'' \emph{Biological cybernetics}, vol. 105, no. 5-6, pp.
  355--370, 2011.

\bibitem{dollar2010highly}
A.~M. Dollar and R.~D. Howe, ``The highly adaptive {SDM} hand: Design and
  performance evaluation,'' \emph{The International Journal of Robotics
  Research}, vol.~29, no.~5, pp. 585--597, 2010.

\bibitem{amend2012positive}
J.~R. Amend, E.~M. Brown, N.~Rodenberg, H.~M. Jaeger, and H.~Lipson, ``A
  positive pressure universal gripper based on the jamming of granular
  material,'' \emph{Robotics, IEEE Transactions on}, vol.~28, no.~2, pp.
  341--350, 2012.

\bibitem{giannaccini2014variable}
M.~E. Giannaccini, I.~Georgilas, I.~Horsfield, B.~Peiris, A.~Lenz, A.~G. Pipe,
  and S.~Dogramadzi, ``A variable compliance, soft gripper,'' \emph{Autonomous
  Robots}, vol.~36, no. 1-2, pp. 93--107, 2014.

\bibitem{Kota2012}
C.~K. Joshua Bishop-Moser, Girish~Krishnan and S.~Kota, ``Design of soft
  robotic actuators using fluid-filled fiber-reinforced elastomeric enclosures
  in parallel combinations,'' in \emph{Proc. {IEEE} ICRA 2012}, Vilamoura,
  Algarve, Portugal, Oct. 7--12, 2012, pp. 4264--4269.

\bibitem{RBOhand2013}
R.~Deimel and O.~Brock, ``A compliant hand based on a novel pneumatic
  actuator,'' in \emph{Proc. {IEEE} ICRA 2013}, Karlsruhe, Germany, May 6--10,
  2013, pp. 2039--2045.

\bibitem{Marchese2014}
C.~D.~O. Andrew D.~Marchese, Konrad~Komorowski and D.~Rus, ``Design and control
  of a soft and continuously deformable 2d robotic manipulation system,'' in
  \emph{Proc. {IEEE} ICRA 2014}, Hong Kong, China, May 31 -- June 7 2014, pp.
  2189--2196.

\bibitem{bilodeau15}
R.~Adam~Bilodeau, E.~L. White, and R.~K. Kramer, ``Monolithic fabrication of
  sensors and actuators in a soft robotic gripper,'' in \emph{Intelligent
  Robots and Systems (IROS), 2015 IEEE/RSJ International Conference on}.\hskip
  1em plus 0.5em minus 0.4em\relax IEEE, 2015, pp. 2324--2329.

\bibitem{homberg15}
B.~S. Homberg, R.~K. Katzschmann, M.~R. Dogar, and D.~Rus, ``Haptic
  identification of objects using a modular soft robotic gripper,'' in
  \emph{Intelligent Robots and Systems (IROS), 2015 IEEE/RSJ International
  Conference on}.\hskip 1em plus 0.5em minus 0.4em\relax IEEE, 2015, pp.
  1698--1705.

\bibitem{YLPark-3axis-eGaIn2012}
Y.-L. Park, B.-R. Chen, and R.~J. Wood, ``Design and fabrication of soft
  artificial skin using embedded microchannels and liquid conductors,''
  \emph{IEEE Sensors Journal}, vol.~12, no.~8, pp. 2711--2718, 2012.

\bibitem{case2015soft}
J.~C. Case, E.~L. White, and R.~K. Kramer, ``Soft material characterization for
  robotic applications,'' \emph{Soft Robotics}, vol.~2, no.~2, pp. 80--87,
  2015.

\bibitem{calli2015ycb}
B.~Calli, A.~Singh, A.~Walsman, S.~Srinivasa, P.~Abbeel, and A.~M. Dollar,
  ``The {YCB} object and model set: Towards common benchmarks for manipulation
  research,'' in \emph{Advanced Robotics (ICAR), 2015 International Conference
  on}.\hskip 1em plus 0.5em minus 0.4em\relax IEEE, 2015, pp. 510--517.

\bibitem{mcevoy2016shape}
M.~McEvoy and N.~Correll, ``Shape change through programmable stiffness,'' in
  \emph{Experimental Robotics}.\hskip 1em plus 0.5em minus 0.4em\relax
  Springer, 2016, pp. 893--907.

\end{thebibliography}

\end{document}